\documentclass[conference]{IEEEtran}
\IEEEoverridecommandlockouts
\usepackage{cite}
\usepackage{amsmath,amssymb,amsfonts}
\usepackage{hyperref}
\hypersetup{
    colorlinks=true,
    linkcolor=blue,
    filecolor=magenta,      
    urlcolor=cyan,
    pdftitle={Overleaf Example},
    pdfpagemode=FullScreen,
    }
\usepackage{graphicx}
\usepackage{textcomp}
\usepackage{xcolor}
\usepackage{booktabs}

\usepackage[capitalize]{cleveref}
\usepackage{caption}
\usepackage{subcaption}
\usepackage{float}
\usepackage[htt]{hyphenat} 
\usepackage{array}
\usepackage{arydshln}

\usepackage{svg}
\usepackage{algorithm}
\usepackage{algpseudocode}
\usepackage{pifont}
\usepackage{lipsum}  
\usepackage{diagbox}
\usepackage{multirow}

\crefname{section}{Sec.}{Secs.}
\Crefname{section}{Section}{Sections}
\Crefname{table}{Table}{Tables}
\crefname{table}{Tab.}{Tabs.}

\def\BibTeX{{\rm B\kern-.05em{\sc i\kern-.025em b}\kern-.08em
    T\kern-.1667em\lower.7ex\hbox{E}\kern-.125emX}}
\begin{document}

\title{Carousel: A High-Resolution Dataset for Multi-Target Automatic Image Cropping}

\author{\IEEEauthorblockN{Rafe Loya,
Andrew Hamara, Benjamin Estell,
Benjamin Kilpatrick, Andrew C. Freeman}
\IEEEauthorblockA{
Baylor University\\
}}







\maketitle

\newcolumntype{v}{>{\sbox0\bgroup}c<{\egroup\raisebox{\dimexpr 0.5\dp0-0.5\ht0}%
 [\dimexpr 0.5\ht0-0.5\dp0+1pt][\dimexpr 0.5\ht0-0.5\dp0+1pt]{\usebox0}}}

\newcommand{\tabcaption}{\refstepcounter{subfigure}(\thesubfigure)}

\begin{abstract}
    Automatic image cropping is a method for maximizing the human-perceived quality of cropped regions in photographs. Although several works have proposed techniques for producing singular crops, little work has addressed the problem of producing multiple, distinct crops with aesthetic appeal. In this paper, we motivate the problem with a discussion on modern social media applications, introduce a dataset of 277 relevant images and human labels, and evaluate the efficacy of several single-crop models with an image partitioning algorithm as a pre-processing step. The dataset is available at \href{https://github.com/RafeLoya/carousel}{https://github.com/RafeLoya/carousel}.

\end{abstract}


\section{Introduction}

Image cropping is a fundamental processing task to remove extraneous pixels from an image. For downstream machine-oriented vision tasks, image cropping needs only to preserve the relevant pixels necessary to maximize the application accuracy. Automatic (or ``aesthetic'') image cropping, on the other hand, is aimed toward \textit{human} vision \cite{ava, aadb}. For this task, the goal is to maximize the perceptual quality of the cropped image based on photographic principles  such as symmetry, the rule of thirds, and leading lines \cite{handcrafted_1, handcrafted_2, handcrafted_3, handcrafted_4, handcrafted_5, handcrafted_6}.

With billions of images uploaded each day, automatic cropping is increasingly compelling for users sharing images on social media. Despite the ever-increasing upload demands, however, these platforms continue to downscale the resolution of all images to save on storage and transmission costs. For example, Instagram supports a maximum horizontal resolution of only 1080 pixels \cite{noauthor_image_2025}. Meanwhile, the sensor resolution in everyday smartphones continues to increase. Indeed, the flagship Samsung Galaxy S25 Ultra has a 200 MP primary sensor \cite{noauthor_galaxy_2025}. If uploaded to Instagram, such an image would thus be downscaled to less than one tenth of the original resolution.

We argue that many photograph categories, such as landscapes, panoramas, and multi-subject candids, are underserved by the severe downscaling introduced by these platforms. While one may zoom in on a published image, the detail from the original high-resolution image is often lost. Inspired by this interactive zoom, we propose \textit{multi-target automatic image cropping}. In this task, we aim to determine the salient regions of an image that a user would likely be most interested in viewing in higher detail. For each region, the cropping algorithm should produce a distinct image that is aesthetically pleasing by itself. Thus, a user receives multiple, distinct image crops from a single input image. These images can then be uploaded as a sequence on a social media platform, allowing viewers to swipe between them to \textit{emulate the experience of dynamically zooming} in on the full-resolution source image.

\begin{figure}
    \centering
    \includegraphics[width=\linewidth]{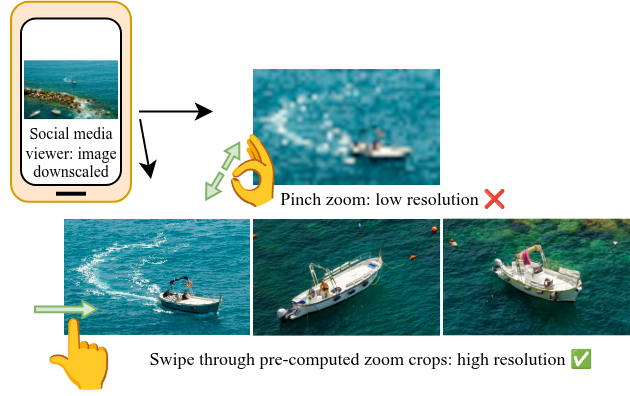}
    \caption{Motivating example of multi-target image cropping.}
    \label{fig:motivation_diagram}
\end{figure}

In this paper, we first examine the literature and discuss the challenges with adapting existing image cropping methods to this task. We then introduce \textit{Carousel}, our dataset of 277 high-resolution multi-target images with human-labeled crops. Finally, we discuss experimental results in evaluating several existing single-target cropping algorithms on our dataset. We conclude that future work should explore methods to produce multi-target crops directly, without the need for significant pre-processing.

\begin{table*}
  \centering
  \begin{tabular}{@{}lc||cccc@{}}
    \toprule
    Year & Paper  & Avg. Resolution & \# Images & Multiscale? & Distinct Crops?  \\ \hline
    2014 & FLMS \cite{fang_automatic_2014} & 0.76 MP & 500 & No & No \\
    2018 & CPC \cite{wei_good_2018} & 0.31 MP & 10,797 & No & No \\
    2019 & GAIC \cite{zeng_reliable_2019} & 0.72 MP & 1,236 & Yes & No \\
    2023 & SACD \cite{yang_focusing_2023} & 0.27 MP & 2,077 & Yes & Sometimes (if multiple subjects)  \\
    2025 & \textbf{Carousel (ours)}  &  10.58 MP & 277 & No & Yes \\
    \bottomrule
  \end{tabular}
  \caption{A comparison of datasets for automatic image cropping. }
  \label{tab:datasets}
\end{table*}

\section{Related Work}

\subsection{Image Cropping Models}

Wei et al. introduced the View Proposal Network (VPN) \cite{wei_good_2018}, which can produce multiple view suggestions with confidence rankings. From the available literature, this method most closely aligns with our task; however, the top-ranked image views are not disjoint, so there is often significant overlap between the cropped images. 

Also utilized by Wei et al., the View Evaluation Network (VEN) \cite{wei_good_2018} served as a teacher model for VPN due to its superior accuracy across multiple benchmarks. The authors considered it too slow for real-time applications, but wished to preserve its performance while increasing the inference speed.

The Aesthetics Aware Reinforcement Learning (A2-RL) framework \cite{a2rl} put forth by Li et al. takes a sequential decision-making approach, where the cropping agent progressively transforms the cropping window through a series of actions until it takes a termination action. It outperforms prior weakly-supervised methods with fewer candidate windows, while having faster inference times.

Compared to VPN, the GAICv2 model \cite{zeng_grid_2020} reduces the number of candidate crops tenfold. This is done through a grid anchor-based approach which considers key aspects such as aspect ratio and content preservation. The result is substantially higher inference speed and accuracy than VPN, VEN, and A2-RL.

The model of Jia et al. \cite{jia_rethinking_2022} also uses an anchoring mechanism to suggest a variety of starting crop regions, then regresses those suggestions to produce aesthetically high-quality crops. This method can produce a wide range of crops at different aspect ratios, but there is no built-in logic to ensure that these crops cover distinct subjects.

The Sac-NET model \cite{yang_focusing_2023} too can produce crops from multiple views. However, this method requires manual user input to first select the subject of interest. The model then restricts the generated crop to be derived from the selected region. In this way, it may be considered a form of \textit{semi-automatic} image cropping.

\subsection{Existing Datasets}

Several datasets have been released for single-view image cropping. Recent works have emphasized multiscale labels for each crop region, typically by providing several aspect ratios (e.g., 2:3, 5:7, etc.) or providing similar crops from a number of human annotators. The SACD dataset used for training Sac-NET \cite{yang_focusing_2023} notably employed an iterative sub-image labeling procedure from multiple annotators, sometimes resulting in distinct subjects being covered by the labels; however, most image labels focus on a single subject. While most datasets source their images from open platforms such as Flickr, they are distributed at very low resolutions. This is acceptable for most single-image cropping methods, because they typically operate on a global scale, discarding very little of the original image. For our multi-target cropping task, however, we are more interested in small subjects. Therefore, a model may require high-resolution detail for compelling performance. We summarize the relevant characteristics of several common datasets in \cref{tab:datasets}.



\section{Our Dataset}

To address the current gap in the literature, we introduce the high-resolution \textit{Carousel} dataset for multi-target automatic image cropping.

\subsection{Sourcing}

The images were sourced from Wikimedia Commons and image aggregators such as Flickr and Rawpixel. All images were distributed with open, non-commercial licenses. We enforced a high-resolution requirement by ensuring that each image was at least 1 MP in size. We only included images with at least two distinct regions of saliency (positive examples) and high aesthetic quality. The average resolution of our images is 10.58 MP, which is an order of magnitude greater than the existing single-target datasets (\cref{tab:datasets}).

\subsection{Labeling}

Our images were sourced and annotated by human experts. 
The reviewers recorded metadata for the license type, the creator of the photograph, the title of the image, the source URL, and the number of distinct salient regions, $k$.

After this reviewing process, the images were annotated with ground truth bounding box labels. We modified AnyLabeling, a popular open-source data annotation platform \cite{nguyen_anylabeling_2025}, to support fixed aspect ratios. 
For this work, we limited our annotations to the 2:3 and 3:2 aspect ratios for landscape and portrait orientations, respectively. Annotators marked the ideal compositions focused on their identified salient regions, referencing classical photography principles such as leading lines, the rule of thirds, and visual balance.


\subsection{Usage}

Our dataset is freely available on GitHub\footnote{\href{https://github.com/RafeLoya/carousel}{https://github.com/RafeLoya/carousel}}. It consists of 277 images. Each image is distributed with a corresponding JSON file containing the ground truth labels and JSON file containing the image metadata (such as the source URL and copyright information). The total dataset size is approximately 1.4 GB. The dataset README provides further details on the file organization and metadata formatting.





\section{Evaluation on Existing Models}

To benchmark current methods against our multi-target cropping task, we tested several popular models with our dataset. Since these models are designed primarily for single-target cropping, we leverage a pre-processing image partitioning step described below.

\subsection{Multi-Region Saliency Partitioning}\label{sec:partitioning}

\begin{figure}
    \centering
    \begin{subfigure}{0.45\linewidth}
        \centering
        \includegraphics[width=\linewidth]{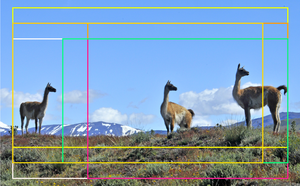}
        \caption{Original image}
        \label{fig:no_partitioning}
    \end{subfigure}
    \hfill
    \begin{subfigure}{0.50\linewidth}
        \centering
        \includegraphics[width=\linewidth]{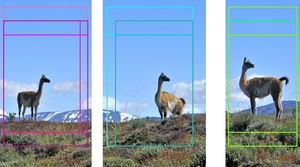}
        \caption{Partitions}
        \label{fig:partitioned}
    \end{subfigure}
    \caption{Comparison of highest-scoring crops from GAICv2 on an image and its partitions, derived from the multi-region saliency partitioning algorithm. Note the significant overlap present in the crops predicted for (a), as the model is designed to produce a single optimized crop.}
    \label{fig:comparative_views}
\end{figure}

To ensure fair evaluation, we enforced non-overlapping crop proposals by partitioning images based on their salient regions. This constraint was necessary because preliminary analysis revealed that existing multi-crop ranking methods produce spatially overlapping proposals, and neglected secondary subjects in favor of dominant ones. For example, ~\cref{fig:no_partitioning} demonstrates how GAICv2 generates overlapping crops that fail to capture the three subjects, predicting only 1-2 distinct crops. ~\cref{fig:partitioned} shows results after partitioning the same image into distinct subregions, ensuring each salient subject receives dedicated cropping attention.

To partition input images into distinct subregions, we leverage the multi-region saliency partitioning algorithm introduced by Hamara et al. \cite{2025croppingalgorithm}. This algorithm segments distinct subjects of interest within each image, enabling the generation of non-overlapping crop regions for downstream processing. We utilized U\textsuperscript{2}-Net \cite{u2net} to generate saliency maps, which are then used for partitioning.

In each iteration (determined by \textit{k}, the number of target crops), we select the smallest valid bounding box that meets a dynamically updated saliency threshold. Once found, the attention values in this region are zeroed-out and the process repeats. Our implementation enhances \cite{2025croppingalgorithm} by determining the partition orientation from the variance in the \textit{x} and \textit{y} positions of the bounding boxes. A greater \textit{x} variance yields vertical partitions, while a greater \textit{y} variance yields horizontal partitions. The space between each pair of adjacent bounding boxes is bisected to produce our partitions.

\subsection{Metrics}

 A common metric in the cropping literature is the Intersection Over Union (IoU), which is given by the spatial area where two bounding boxes overlap (intersection), divided by the total area of the image they cover (union). Whereas many models will evaluate IoU only for the crop with the highest confidence, the A2-RL model does not provide confidence scores. Furthermore, our dataset provides multiple ground truth labels for each image, so we require a method for matching cropped bounding boxes to the appropriate labels.

 Therefore, we define the metric Top-$k$ IoU (or ``kIoU'') and our evaluation methodology as follows. For each of the $k$ partitions for an image, we first transform the coordinates of the inference bounding boxes back to the original image space. We then calculate the intersection over union (IoU) of each combination of predicted and ground truth bounding boxes. For a given IoU threshold, $\mu$, we begin by identifying all prediction-ground truth pairs that exceed this threshold. Next, we sort these pairs by their IoU score in descending order. Finally, we perform greedy bipartite matching, prioritizing the pair with the highest IoU score; once a pair is matched, both bounding boxes are removed from further consideration. This process continues $k$ times, until no valid pairs remain for the image. Across all images in the dataset, we average the resulting IoU pairs to yield the kIoU. Specifically, we calculate kIoU@0.5 (with $\mu=$ 50\%) and kIoU@0.5:0.95 (averaging the kIoU values for $\mu \in \{0.5, 0.55, 0.6, \ldots, 0.95 \}$).


%


\begin{table}
    \centering
    \begin{tabular}{c|ll}
        \toprule
        Model & kIoU@0.5:0.95 & kIoU@0.5 \\ \hline 
        VPN* \cite{wei_good_2018} & 0.068 & 0.179 \\
        VPN \cite{wei_good_2018} & 0.223 & 0.565 \\
        A2RL \cite{a2rl} & 0.145 & 0.409 \\
        VEN \cite{wei_good_2018} & 0.210 & 0.538 \\
        GAICv2 \cite{zeng_grid_2020} & \textbf{0.231} & \textbf{0.574} \\
        \bottomrule
    \end{tabular}
    \caption{Quantitative performance of existing aesthetic cropping models on our dataset. VPN* uses the top multi-view outputs from the VPN model on the original images, while the other models listed use our multi-region saliency partitioning algorithm (\cref{sec:partitioning}) as a pre-processing step. }
    \label{tab:map_results}
\end{table}

\begin{figure}[h]
    \centering
    \begin{subfigure}{0.45\linewidth}
        \centering
        \includegraphics[width=\linewidth]{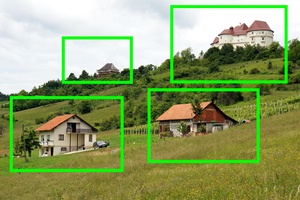}
        \caption{Ground truth labels}
        \label{fig:croatia_gt_labels}
    \end{subfigure}
    \begin{subfigure}{0.45\linewidth}
        \centering
        \includegraphics[width=\linewidth]{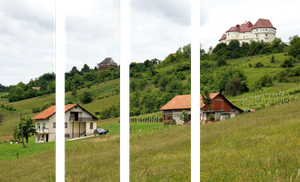}
        \caption{Partitions}
        \label{fig:croatia_partitioned}
    \end{subfigure}
    \caption{Example failure case for our partitioning algorithm.}
    \label{fig:comparative_views}
\end{figure}

\newcommand{\imwidth}{0.86\linewidth}

\begin{figure*}[!t]
    \centering
    \begin{tabular}{m{0.19\textwidth}m{0.16\textwidth}m{0.16\textwidth}m{0.08\textwidth}m{0.16\textwidth}}
        (a) Ground truth labels 
            & \includegraphics[width=\imwidth]{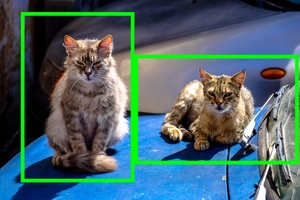}
            & \includegraphics[width=\imwidth]{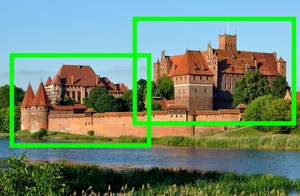}
            & \includegraphics[width=\imwidth]{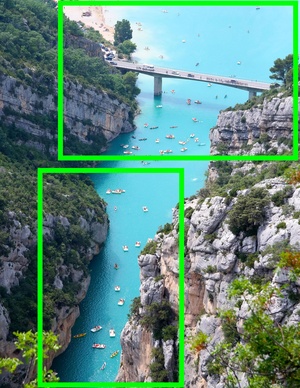}
            & \includegraphics[width=\imwidth]{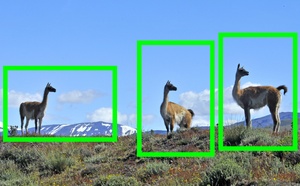} \\
        (b) VPN* \cite{wei_good_2018}
            & \includegraphics[width=\imwidth]{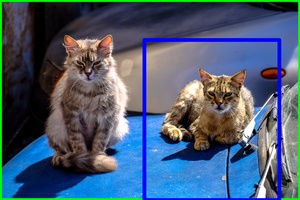}
            & \includegraphics[width=\imwidth]{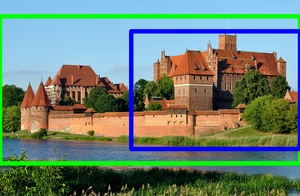}
            & \includegraphics[width=\imwidth]{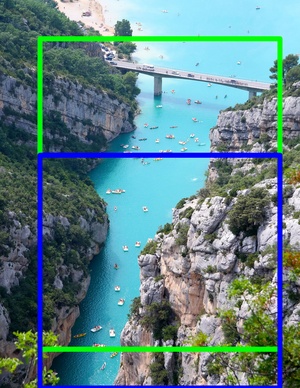}
            & \includegraphics[width=\imwidth]{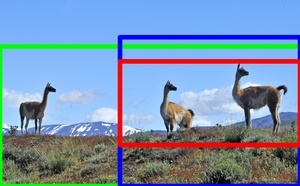} \\
        (c) VPN \cite{wei_good_2018}
            & \includegraphics[width=\imwidth]{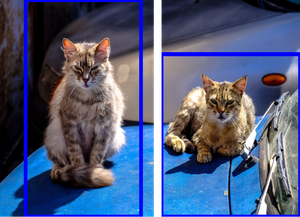}
            & \includegraphics[width=\imwidth]{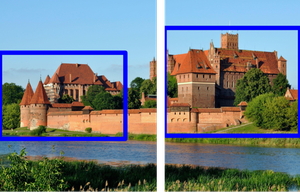}
            & \includegraphics[width=\imwidth]{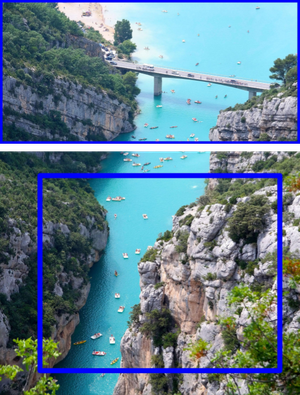}
            & \includegraphics[width=\imwidth]{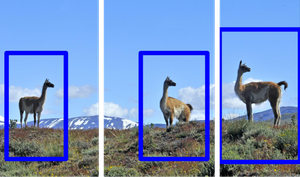} \\
        (d) A2-RL \cite{a2rl}
            & \includegraphics[width=\imwidth]{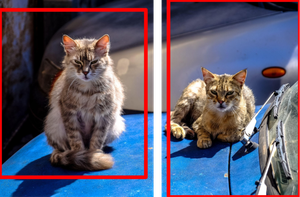}
            & \includegraphics[width=\imwidth]{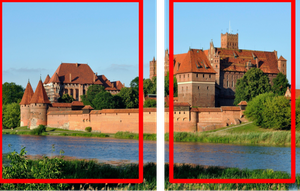}
            & \includegraphics[width=\imwidth]{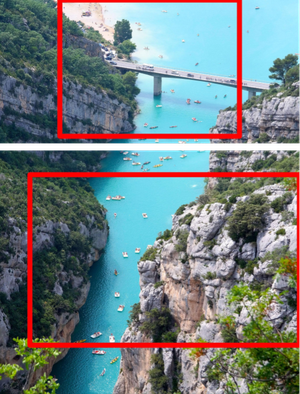}
            & \includegraphics[width=\imwidth]{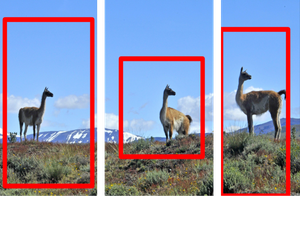} \\
        (e) VEN \cite{wei_good_2018}
            & \includegraphics[width=\imwidth]{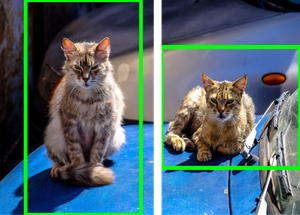}
            & \includegraphics[width=\imwidth]{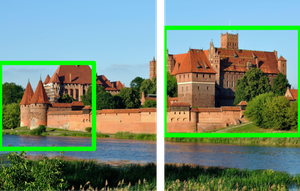}
            & \includegraphics[width=\imwidth]{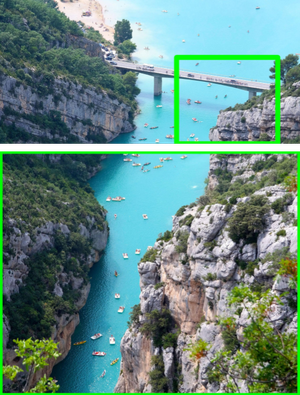}
            & \includegraphics[width=\imwidth]{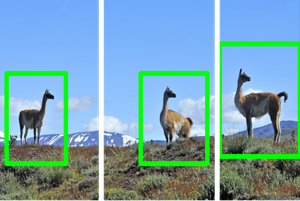} \\
        (f) GAICv2 \cite{zeng_grid_2020}
            & \includegraphics[width=\imwidth]{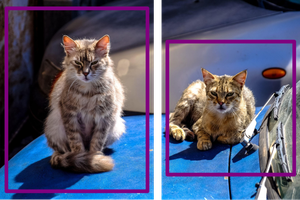}
            & \includegraphics[width=\imwidth]{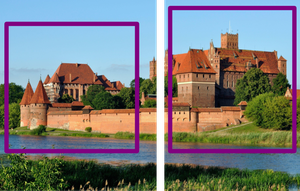}
            & \includegraphics[width=\imwidth]{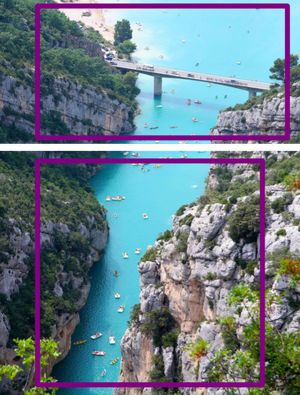}
            & \includegraphics[width=\imwidth]{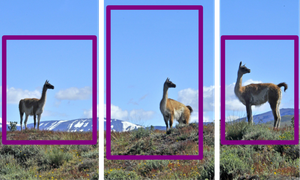} \\
    \end{tabular}
    \caption{Visual comparison of multi-target crops on our dataset. (b) shows the multi-view outputs of VPN on the original images, while (c)-(f) use our multi-region saliency partitioning algorithm (\cref{sec:partitioning}) followed by the single-target cropping models. 
    } 
    \label{fig:model_comparison}
\end{figure*}

\subsection{Results}





232 of the 277 images were considered when evaluating performance on the partitioned dataset. The remaining 45 photographs were omitted due to failures that occurred during the partitioning pre-processing step. Failure in this context was defined as partitions that approximately bisect one or more ground truth labels, and those with extreme aspect ratios (e.g., extremely narrow)


For example, \cref{fig:croatia_gt_labels} illustrates a difficult set of ground truth labels, where the crops have significant coordinate overlap in both $x$- and $y$-dimensions. The partitioning algorithm assumes that all image divisions will have the same orientation, so the resulting partitions (\cref{fig:croatia_partitioned}) make reasonable crops impossible. Similar failures occur when there is a strong disparity between the spatial size of different salient regions.



For a fair comparison of the cropping algorithms, we evaluated the remaining 232 images as follows. Using our image partitioning step and the ground-truth counts $k$ for the number of target crops, we tested the View Proposal Net (VPN) \cite{wei_good_2018}, A2-RL \cite{a2rl}, the View Evaluation Net (VEN) \cite{wei_good_2018}, and GAICv2 \cite{zeng_grid_2020} models. We further evaluate VPN on the un-partitioned images, extracting the top \textit{k} view suggestions.

We summarize our kIoU accuracy results in \cref{tab:map_results}. Through the use of our multi-target partitioning step, we see that the GAICv2 model performed the best in our multi-target cropping evaluation. This result is consistent with prior IoU results for single-target cropping\cite{zeng_grid_2020}, where GAICv2 outperforms the other models across many datasets. We further see that the VPN model with its multi-view selection on the original images (denoted VPN* in \cref{tab:map_results}) performs far worse than all models that first leverage our partitioning step. 

The qualitative results in \cref{fig:model_comparison} underscore this point, as we see that VPN* frequently produces crops with high levels of overlap. \cref{fig:comparative_views} further shows that GAICv2's default suggested crops, without partitioning, fail to consider that the user may want to emphasize the three distinct llama subjects. In contrast, with partitioning, the emphasis on distinct subjects becomes achievable, as shown in \cref{fig:model_comparison}.

Although these results point to the efficacy of partitioning for adapting single-target models to the multi-target domain, we emphasize that further work will be necessary to achieve useful results in more difficult scenarios of our dataset (the partitioning ``failure'' cases).


\section{Future Work}

Although our partitioning algorithm provides a compelling interface for traditional single-target cropping models, it can be a weak point when the derived partitions are poor quality or if the ideal compositions are not completely disjoint. We include in our dataset the 45 images for which partitioning ``failed,'' to encourage further research. Future work should focus on designing a novel model that can produce the multi-target image crops \textit{directly} from the full-resolution source image. Such methods should also automatically determine the ideal number of target crops ($k$). It may be helpful to extend our dataset to include more diverse subject categories and ground truth labels in additional aspect ratios.

\section{Conclusion}

We present \textit{Carousel}, a dataset with 277 high-resolution images containing multiple distinct subcrops. By recording the number of target crops for a given photograph, the multi-region saliency partitioning algorithm can be utilized to divide an image to be processed by common single-target cropping models. While this method is promising, this dataset unlocks the potential for future cropping models to output distinct images directly. These efforts may lead to more interactive and high-quality experiences for social media users, especially when viewing high-resolution photos.

\bibliographystyle{IEEEtran}
\bibliography{references0, references2} 

\end{document}